\documentclass[10pt,twocolumn,letterpaper]{article}

\usepackage{iccv}
\usepackage{times}
\usepackage{epsfig}
\usepackage{graphicx}
\usepackage{amsmath}
\usepackage{amssymb}

\usepackage[pagebackref=true,breaklinks=true,letterpaper=true,colorlinks,bookmarks=false]{hyperref}

\iccvfinalcopy 

\def\iccvPaperID{8683} 

\ificcvfinal\fi

\begin{document}

\title{DualMix: Unleashing the Potential of Data Augmentation for Online Class-Incremental Learning}


\author{%
   Yunfeng~Fan$^1$,
   Wenchao~Xu$^1$,
   Haozhao~Wang$^{2,}$\thanks{Corresponding author},
   Jiaqi Zhu$^3$,
   Junxiao~Wang$^{4,5}$,
   and~Song~Guo$^1$\\
  \textsuperscript{1}The Hong Kong Polytechnic University,
  \textsuperscript{2}Huazhong University of Science and Technology\\
  \textsuperscript{3}School of Automation, Beijing Institute of Technology,
  \textsuperscript{4}KAUST, \textsuperscript{5}SDAIA-KAUST AI Center\\
  \texttt{ $\left\{  \right. $yunfeng.fan,wenchao.xu$\left.  \right\} $@polyu.edu.hk}, \texttt{hz\_wang@hust.edu.cn}, \\
\texttt{E1111838@u.nus.edu}, \texttt{junxiao.wang@kaust.edu.sa}, \texttt{song.guo@polyu.edu.hk}
}


\maketitle
\ificcvfinal\thispagestyle{empty}\fi

\begin{abstract}
Online Class-Incremental (OCI) learning has sparked new approaches to expand the previously trained model knowledge from sequentially arriving data streams with new classes. Unfortunately, OCI learning can suffer from catastrophic forgetting (CF) as the decision boundaries for old classes can become inaccurate when perturbated by new ones. Existing literature have applied the data augmentation (DA) to alleviate the model forgetting, while the role of DA in OCI has not been well understood so far. 
%
In this paper, we theoretically show that augmented samples with lower correlation to the original data are more effective in preventing forgetting. However, aggressive augmentation may also reduce the consistency between data and corresponding labels, which motivates us to exploit proper DA to boost the OCI performance and prevent the CF problem. We propose the Enhanced Mixup (EnMix) method that mixes the augmented samples and their labels simultaneously, which is shown to enhance the sample diversity while maintaining strong consistency with corresponding labels. Further, to solve the class imbalance problem, we design an Adaptive Mixup (AdpMix) method to calibrate the decision boundaries by mixing samples from both old and new classes and dynamically adjusting the label mixing ratio. Our approach is demonstrated to be effective on several benchmark datasets through extensive experiments, and it is shown to be compatible with other replay-based techniques.

\end{abstract}

\section{Introduction}
\label{sec: intro}

Deep learning (DL) has made remarkable achievements by imitating human intelligence to mine the knowledge from carefully gathered datasets. Further inspired by human learning process, continual learning (CL), also known as incremental learning, has taken a next step to broaden the model knowledge from a number of sequentially arrived tasks \cite{lopez2017gradient, shin2017continual, van2019three}. However, while learning new tasks, CL can suffer from \textit{catastrophic forgetting} (CF) that the model’s classification capability can drop severely on old tasks \cite{mccloskey1989catastrophic, goodfellow2013empirical, kirkpatrick2017overcoming}. The CF phenomenon can be significant especially under more realistic settings, i.e., the online class-incremental (OCI) learning, where new classes are continually arriving with the data stream and every batch samples can be observed only once \cite{he2020incremental, masana2022class}.

\begin{figure}
    \centering
    \includegraphics[width=1.0\linewidth]{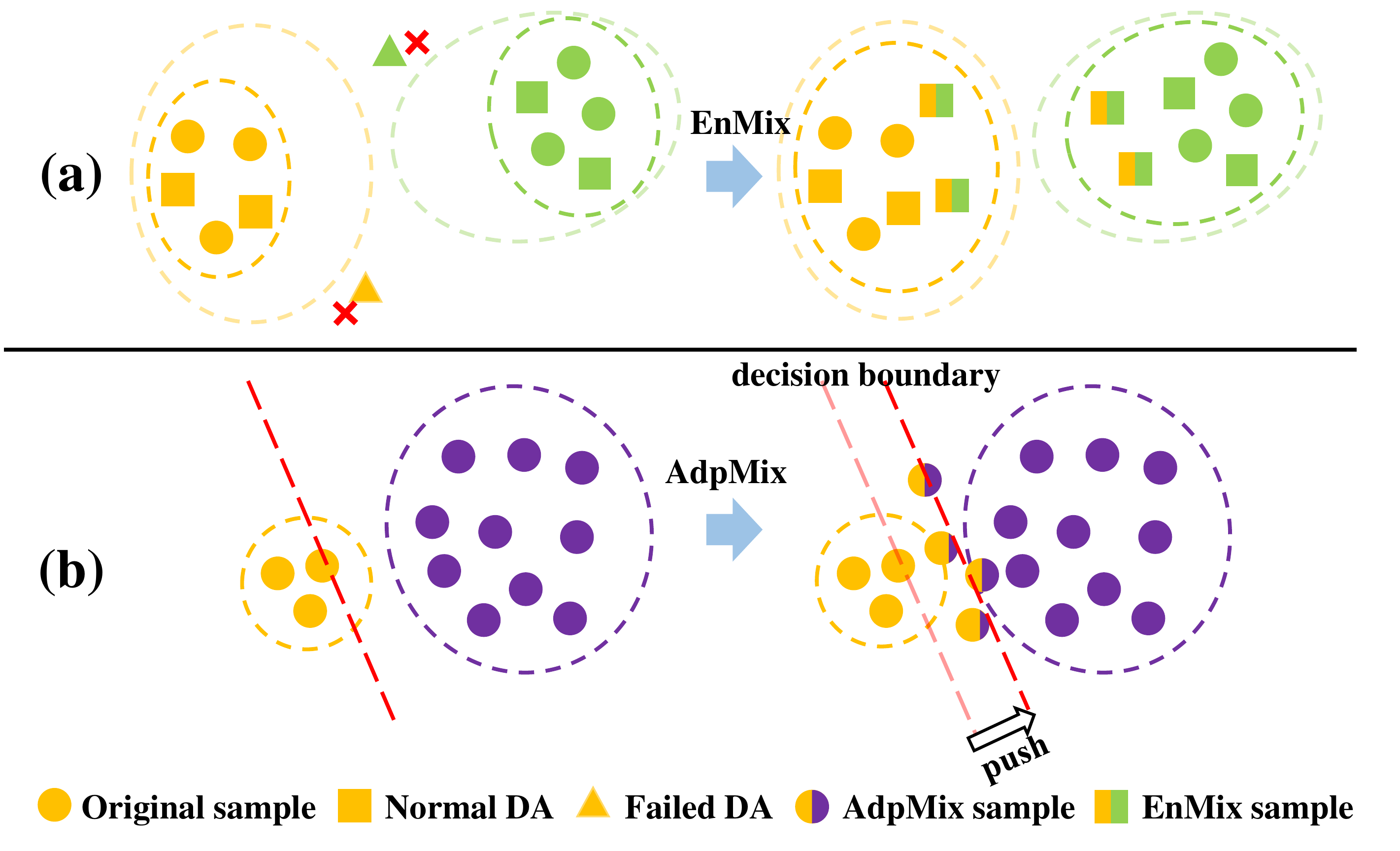}
    \caption{(a) Normal data augmentation increases the diversity of limited samples (within the opaque border), but excessive augmentation can produce samples that are unrelated to the labels (outside of the transparent border). To achieve more robust augmentation, we mix the augmented samples from various classes. (b) In OCI, imbalanced class distribution leads to a biased decision boundary that favors the old classes stored in memory. To overcome this, we mix the samples from both old and new classes and adjust the label mixing ratio to push the decision boundary away and make it more adaptive.}
    \label{fig: mix seperate}
\end{figure}
Replay-based methods \cite{buzzega2021rethinking,mai2021supervised,shim2021online} have been shown to be effective against CF for OCI by storing past task samples in memory buffer and replaying them while training for new tasks. 
Recent research \cite{zhu2021class, zhu2021prototype, mai2021supervised, guo2022online} has aimed to alleviate CF by performing data augmentation (DA) \cite{van2001art, shorten2019survey} to both previously buffered and newly arrived samples.
%
These methods focus on complex loss design coupled with contrastive or self-supervised learning (SSL) \cite{zbontar2021barlow, noroozi2018boosting} techniques to replace the less effective cross-entropy (CE) loss with advanced loss functions such as InfoNCE loss \cite{he2020momentum} or mutual information \cite{gu2022not}.
However, the role of DA in OCI still remains unclear and requires exploitation. 
Different with literature status quo, in this paper, we will consider the input side of OCI, exploring \textit{what constitutes a good DA in OCI and how to apply it effectively?}



\begin{figure}
    \centering
    \includegraphics[width=0.85\linewidth]{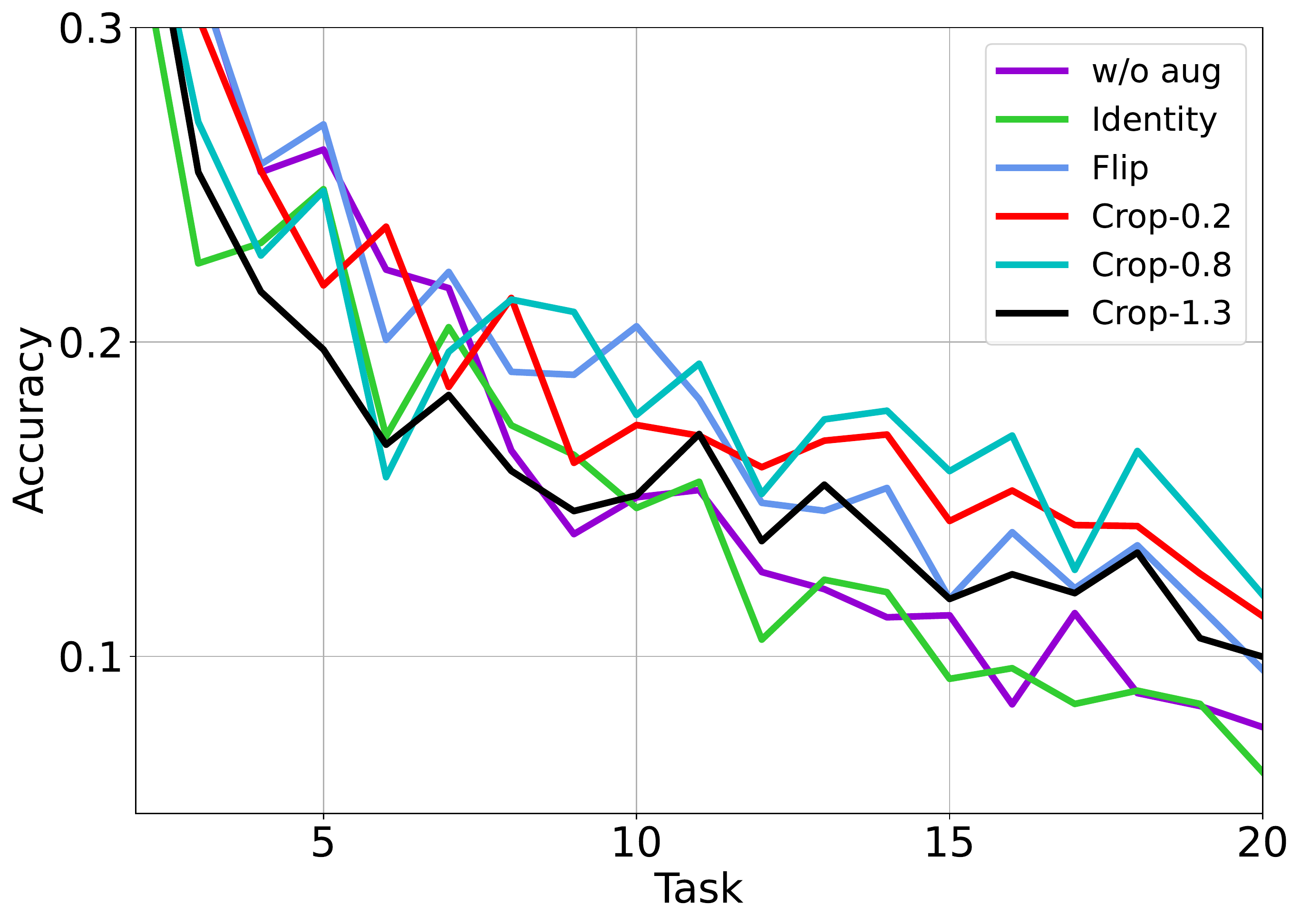}
    \caption{Average accuracy on different augmentation strategies with the baseline ER. ``Identity": copy the original samples as augmentation; ``Flip": random horizontal flip; ``Crop-xx": random resized crop, number indicates strength. ``Identity" is slightly weaker than ER with no augmentation. ``Crop" is usually better and different strengths correspond to different promotions. All experiment were performed on CIFAR-100 and memory size is 1k.}
    \label{fig: aug comparison}
\end{figure}

Intuitively, DA could create samples with extended diversity, which is shown to be beneficial to representation learning \cite{perez2017effectiveness, fruhwirth1994data, wong2016understanding} and can potentially prevent CF in OCI.
However, counterintuitively, \cite{zhang2022simple} reveals that directly applying DA, without multiple iterations \cite{aljundi2019online} on experience replay (ER) \cite{chaudhry2019tiny}, leads to worse performance than ER without augmentation.
Such anomaly inspires us to compare different DA methods for OCI. 
%
As shown in Figure~\ref{fig: aug comparison}, DA does not always harm OCI performance, but selecting the appropriate augmentation method and strength\footnote{For the RandomResizedCrop operator with the scale range $\left[ a,b \right]$, its strength is defined as $\left( 1-a \right) +\left( 1-b \right) $.} is critical.
The results indicate that aggressive augmentation is more likely to perform well in OCI, whereas excessively strong DA can lead to worse performance (Crop-1.3 is worse than Crop-0.8).

%
%

To explain the aforementioned phenomenon, we conducted a theoretical analysis on the relationship between DA and the forgetting in OCI.
Our analysis suggests that the augmented samples from the memory buffer should have low covariance on the model's mean cross entropy compared to the original samples.
However, excessive augmentation with low covariance may generate new samples that deviate from the ground-truth labels, leading to the introduction of erroneous information.
To address this issue, we propose the \textbf{En}hanced \textbf{Mix}up (EnMix), which applies mixup \cite{zhang2017mixup} on augmented samples from memory to ensure stronger DA while still maintaining high consistency with their labels, as shown in Figure \ref{fig: mix seperate}.
Additionally, we observed that the decision boundaries between old and new classes are biased towards old classes due to class imbalance \cite{kim2020imbalanced, cha2021co2l} in OCI. 
To address this, we propose \textbf{Ad}a\textbf{p}tive \textbf{Mix}up (AdpMix) to adjust the decision boundaries between old and new classes according to the weight imbalance from the classifier. The two methods we propose are collectively referred to as DualMix.

%
To sum up, this paper makes the following contributions.
\begin{itemize}
    \item We provide a theoretical explanation that low covariance between DA samples and original samples in terms of mean cross entropy is advantageous in OCI.
    In addition, we introduce EnMix, which enhances standard DA by generating stronger and more dependable samples.

    \item The AdpMix method is proposed to strike a balance between the decision boundaries of old and new classes by dynamically adjusting the mixing ratio of labels.

    \item Our proposed augmentation methods, which only adjust the input side, are shown to outperform existing methods on several benchmark datasets and be compatible with other replay-based techniques, as demonstrated by extensive empirical results.
\end{itemize}

\section{Related works}
\label{sec: related}
\subsection{Replay-based online continual learning}
In the online continual learning (OCL) \cite{cai2021online, liu2020learning} setting, data arrives in small batches sequentially, and previous batches from current and previous tasks cannot be reused, presenting the challenge of efficient single-pass learning.
OCL can be classified into two types based on whether the new task contains new classes: online class-incremental (OCI) and online domain-incremental (ODI) \cite{alfarra2022simcs, gunasekara2022adaptive}. 
This work mainly focuses on the more challenging OCI setting, which relies heavily on replay-based methods.
Chaudhry \etal~\cite{chaudhry2019tiny} proposed experience replay (ER) to store a subset of data from previously seen tasks in memory and replay them during training on new tasks.
Variants of ER have been developed to optimize sample selection and representation learning strategies.
A-GEM \cite{chaudhry2018efficient} constrains the gradient from memory samples to prevent the average loss from previous tasks from increasing.
MIR \cite{NEURIPS2019_15825aee} compares the loss increments of memory samples after updating the model based on current batch data to perform sample retrieval.
GSS \cite{aljundi2019gradient} stores samples with more diversity in gradient directions in memory.
ASER \cite{shim2021online} scores each sample by the Sharply value according to its ability to preserve the latent decision boundaries of previously seen classes.
Although these methods aim to prevent forgetting by storing and revisiting representative memory samples, they do not address the severe class imbalance problem in replay-based OCI. 
GDumb \cite{prabhu2020gdumb} was developed specifically to address class imbalance problem by greedily keeping the number of samples from each class balanced in memory and training the model with only memory data.
However, it excessively reduces the usage of new task data, resulting in insufficient feature exploitation.
In this work, we propose two simple yet efficient methods to prevent forgetting and alleviate class imbalance simultaneously.
Our methods are specifically designed for the OCI setting and only adjust the input side. 
We demonstrate the superiority of our methods on multiple benchmark datasets.

\subsection{Data augmentation and continual learning}
To alleviate the CF phenomenon, various recent studies \cite{mai2021supervised, zhu2021class, guo2022online} have focused on learning representations with appropriate features.
For instance, Mai \etal~\cite{mai2021supervised} proposed SCR, which utilizes contrastive learning to encourage clustering samples from the same class, and replaces the linear classifier with the Nearest-Class-Mean classifier to address the problem of imbalanced weights.
Gu \etal \cite{gu2022not} analyzed representation learning with mutual information and proposed DVC to retain more information from previous tasks, the similar way with \cite{guo2022online}.
Zhu \etal~\cite{zhu2021class} employed data mixture \cite{zhang2017mixup} to generate samples with virtual new classes and was also inspired by SSL to augment classes by rotating training samples to promote learning of holistic features \cite{zhu2021prototype}.
Although these methods use DA to generate diverse samples, their purpose is to satisfy the requirements for contrastive learning or SSL, and they have not examined the role of DA in CL.
Moreover, these methods necessitate changes in the model structure (such as additional classifier heads for new classes) during training. 
In this paper, we investigate the role of DA in replay-based OCI with theoretical explanations, and propose two DA methods that solely regulate the input side.

%
%
%
%
%


\section{Method}
\label{sec: method}
\subsection{Insights derived from theoretical exposition}
\label{sec: insight}
In this paper, we study a supervised OCI learning setting as in \cite{mai2022online, aljundi2019online, lin2022anchor}. Consider a data stream $D=\left\{ D_1,D_2,...,D_N \right\} $ and $D_i=\left( X_i,Y_i \right)$ is the data for task $i$. $X_i$ and $Y_i$ represent the samples and corresponding labels in task $i$. $Y_i\cap Y_j=\emptyset $ for $i\ne j$. In the training phase, the data $D_i$ from the task $i$ can be seen only once, which means only one epoch for training is allowed. The model can be divided into feature extractor $f_{\theta}$, with parameters $\theta$, and a linear classifier (single-head for all classes) with weights $\left\{ \mathbf{w}_c \right\} _{c=1}^{C}$ ($C$ is the class number for all tasks, and we omit the bias for simplification). 
We denote $\mathbf{h}_{i}=f_{\theta}\left( x_{i} \right) \in \mathcal{R} ^d $ as the extracted features of sample $x_i$. $F_{\theta}\left( x_{i} \right)=\sigma \left( \mathbf{w}^Tf_{\theta}\left( x_{i,j} \right) \right) \in \mathcal{R} ^C$ is the output probabilities of sample $x_i$. $\sigma \left( \cdot \right)$ is the softmax operation.

When the model is trained on task $t$, according to \cite{gu2022not}, the empirical risk which the model aims to minimize is defined as 
\begin{equation}
    \begin{split}
        R_t\left( F \right) \overset{def}{=}&\underset{\left( x,y \right) \sim D_t}{\mathrm{E}}\left[ L\left( y,F\left( x \right) \right) \right] + \\
        &\beta \lambda \underset{\left( x,y \right) \sim D_{t-1}^{\mathcal{M}}}{\mathrm{E}}\left[ L\left( y,F\left( x \right) \right) \right] 
    \end{split} 
    \label{eq: risk_no_aug}
\end{equation}
where $D_{t-1}^{\mathcal{M}}$ is the fixed-size memory after trained on task $t-1$. We omit $\theta$ in $F_{\theta}$ for simplicity. $\lambda :=\frac{\left| D_t \right|}{\left| D^{\mathcal{M}} \right|} $ and $\beta :=1/\left( 1+\frac{2\left| D_t \right|}{\left| D_{\left[ 1,t \right]} \right|} \right)$. $D_{\left[ 1,t \right]}$ is all the seen data $\left\{ D_1,D_2,...,D_t \right\}$. 
%
%
If we apply a random transform $g\in G$ on memory samples, the expansion memory data is $D^{\mathcal{M} g}=\left\{ x,g\left( x \right) |x\in D^{\mathcal{M}} \right\} $.
The risk with DA of the second term in Equation~\ref{eq: risk_no_aug}, indicating the accquired knowledge from past tasks, is as:
\begin{equation}
    R_{t}^{\mathcal{M} g}\left( F \right) =\beta \lambda \underset{\left( x,y \right) \sim D_{t-1}^{\mathcal{M} g}}{\mathrm{E}}\left[ L\left( y,F\left( x \right) \right) \right] 
    \label{eq: mem_aug risk}
\end{equation}

The objective of CL is to minimize the empirical risk over all the tasks seen so far, so the empirical risk from task 1 to $t-1$ without forgetting should be
\begin{equation}
    R_{t-1}^{obj}\left( F \right) \overset{def}{=}\underset{\left( x,y \right) \sim D_{\left[ 1,t-1 \right]}}{\mathrm{E}}\left[ L\left( y,F\left( x \right) \right) \right] 
    \label{eq: CL obj}
\end{equation}

Therefore, we define the forgetting gap as:
\begin{equation}
    FG^{\mathcal{M} g}=\underset{g}{\mathrm{E}}\left[ \left( R_{t}^{\mathcal{M} g}\left( F \right) -R_{t-1}^{obj}\left( F \right) \right) ^2 \right] 
    \label{eq: forgetting gap}
\end{equation}

We further define the mean covariance of sample's CE loss in dataset $D_{t-1}^{\mathcal{M} g_{}}$ as
\begin{equation}
    CO^{\mathcal{M} g}=\underset{x_i,x_j\sim D_{t-1}^{\mathcal{M} g_{}}}{\mathrm{E}}\left[ \mathrm{Cov}\left[ q\left( x_i \right) ,q\left( x_j \right) \right] \right] 
    \label{eq: true covariance}
\end{equation}
where $\mathrm{Cov}\left[ \cdot ,\cdot \right]$ stands for the covariance. $q$ is defined as $q\left( x_i \right) =-{y_i}^T\log \left( F\left( x_i \right) \right) $.
%

Figure \ref{fig: aug comparison} shows that if DA could generate more diverse information, then the forgetting problem can be alleviated accordingly.
Inspired from~\cite{wangmakes}, we give the rigorous analysis and prove it in the Appendix.
\begin{figure*}[h]
    \centering
    \includegraphics[width=0.8\linewidth]{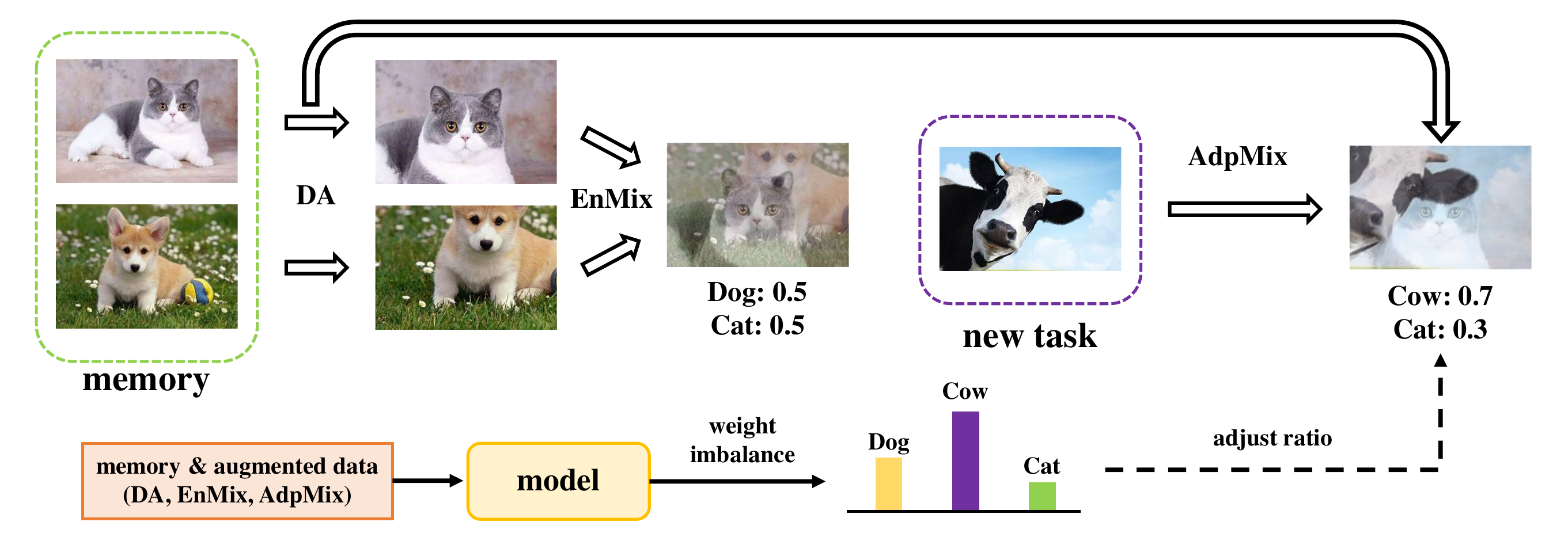}
    \caption{The workflow for integrating EnMix and AdpMix techniques into the learning process of OCI.}
    \label{fig: pipeline}
\end{figure*}

\noindent\textbf{Proposition 1.} Given a model trained on sequential data $\left\{ D_1,D_2,...,D_{t-1} \right\}$ with memory buffer $D^{\mathcal{M}}_{t-1}$, there comes a new task with $D_t$. Considering the data transforms $g_1\in G$ and $g_2\in G$, they are applied on memory samples to obtain $D_{t-1}^{\mathcal{M} g_1}$ and $D_{t-1}^{\mathcal{M} g_2}$. If the mean covariance of CE loss in $D_{t}^{\mathcal{M} g_1} $ is lower than that in $D_{t}^{\mathcal{M} g_2}$, i.e., $CO^{\mathcal{M} g_1}<CO^{\mathcal{M} g_2}$, model will suffer less forgetting on previous tasks, $FG^{\mathcal{M} g_1}<FG^{\mathcal{M} g_2}$.

This proposition shows that DA should reduce the correlation between the augmented and the original samples. Since the covariance is determined by the model output and less forgetting can be achieved if model changes slightly after the new task learning, we calculate the variance of the output probability based on old model as an indicator for evaluating the strength of DA, inspired from \cite{wangmakes}:
\begin{equation}
    \bar{m}=\frac{1}{C_{old}}\sum_{i\in \left[ C_{old} \right]}{\left( \mathrm{Var}_{\mathcal{M} g}\left( \mathbf{u} \right) \right)}^{\frac{1}{2}}, \mathbf{u}=\frac{1}{\left| \mathcal{M} g \right|}\sum_{x_i\in \mathcal{M} g}{F\left( x_i \right)}
    \label{eq: corr calculation}
\end{equation}
See Section~\ref{sec: ablation} for the experimental results.


\subsection{Enhanced mixup augmentation}
As described in Section \ref{sec: insight}, we should conduct a strong DA in the memory for OCI to alleviate forgetting. However, when the DA is too strong, i.e., the output probability of augmented view is far from that of the original sample, the augmented view is likely to keep weak correlation with the label, which will destroy the supervised learning (e.g. ``Crop-1.3" with stronger DA but poor performance than ``Crop-0.8" in Figure \ref{fig: aug comparison}). \textit{How can we further reduce the correlation between the augmented views and the original samples without destroying the correspondence with their labels?} 

In order to solve the problem, we propose a simple yet effective method, enhanced mixup (EnMix), based on intensive DA with reliable label correlation. Given a data transform $g\in G$ and a sample $x$, the augmented view will be $\tilde{x}=g\left( x \right)$. EnMix constructs virtual samples by mixing the augmented views from different original samples and also linearly interpolating their labels:
\begin{equation}
    \begin{aligned}
        & \tilde{x}^e=\mu \tilde{x}_i+\left( 1-\mu \right) \tilde{x}_j \\
        & \tilde{y}^e=\mu y_i+\left( 1-\mu \right) y_j
    \end{aligned}
    \label{eq: enmix}
\end{equation}
where $\mu \in \left[ 0,1 \right]$ and $\mu \sim \mathrm{Beta}\left( \alpha ,\alpha \right)$, $\alpha \in \left( 0,\infty \right) $. Due to mixing with other samples (including other classes) in memory, the model output probability of the mixed samples should be dissimilar with the original samples'. What counts is that the label is also mixed, building correlation between the enhanced samples with different labels. Through this way, we not only reduce the correlation between the augmented views and original samples, but also preserve the consistency between views and labels. The workflow about how to mix is shown as Figure \ref{fig: pipeline}.

\subsection{Adaptive mixup for balance}
In the above sections, we try to produce new samples through DA, enriching the diversity of previous task data in memory and alleviating the CF problem. However, according to previous works~\cite{mai2022online}, the CF phenomenon emerges not only because of the forgetting about previous knowledge, but also due to the inherent class imbalance property in replay-based OCI. The number of samples from new classes is generally greater than the number of samples from old classes in memory, and this imbalance intensifies as the number of tasks increases. 

\begin{figure}
    \centering
    \includegraphics[width=0.8\linewidth]{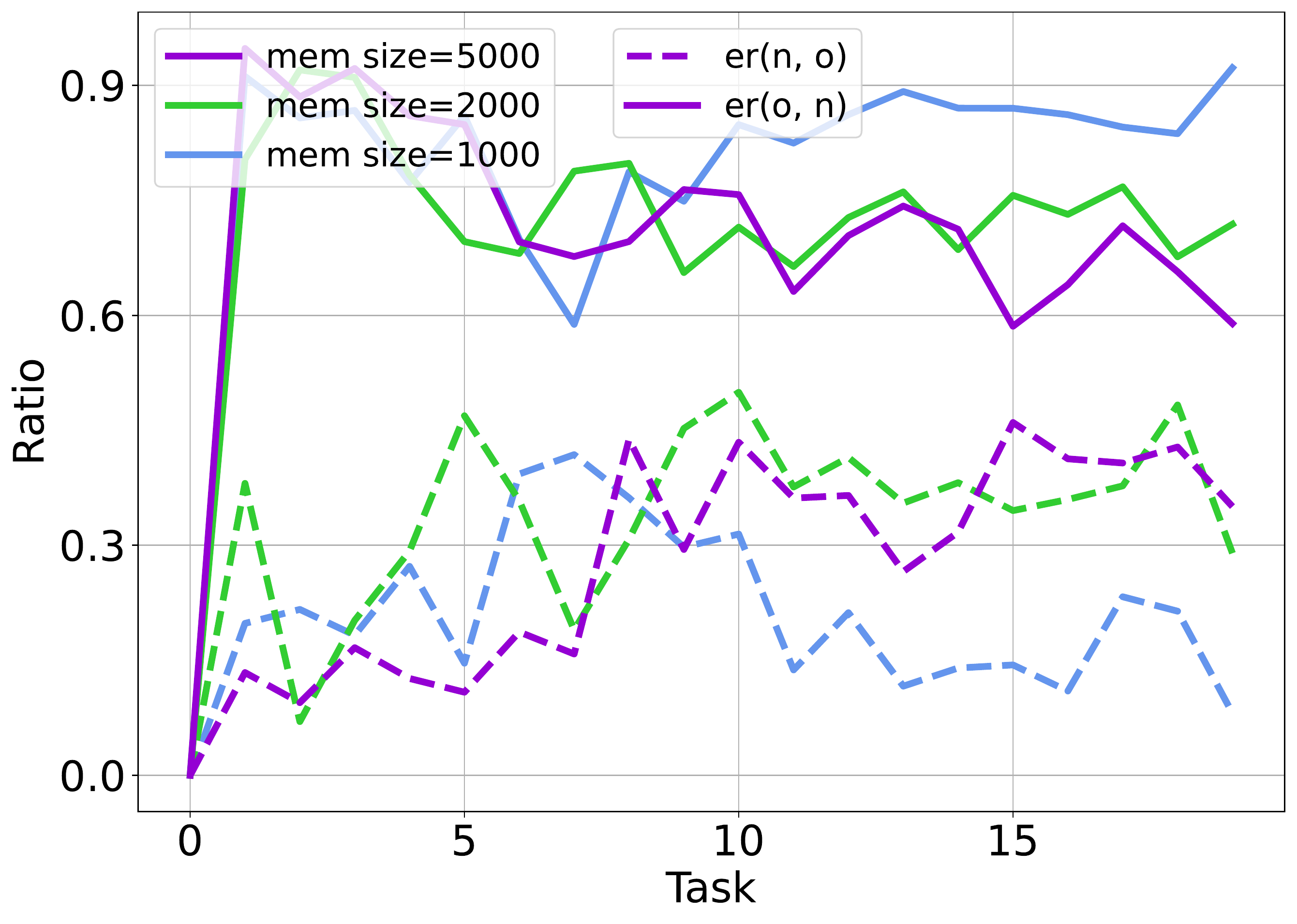}
    \caption{The misclassification ratios during the training process. The results were acquired from CIFAR-100 and trained by ER with different memory size.}
    \label{fig: ER-ratio}
\end{figure}

The class imbalance eventually results in severely biased decision boundary towards the old classes. In order to verify it, we denote the misclassification ratio following \cite{mai2022online}: $er\left( n,o \right)$ denotes the ratio of new class test samples misclassified as old classes to the total number of misclassified new class samples. Same notation rule is applied for $er\left( o,n \right)$. As shown in Figure \ref{fig: ER-ratio}, the samples from old classes can be incorrectly classified to new classes with high probability and the probability of samples from new classes being misclassified to old is relatively small, indicating that the decision boundaries between old and new classes are severely uneven. Moreover, as the memory size decreases, this imbalance is further aggravated. 
In order to adjust the decision boundaries, DA can also be a straightforward method. Performing DA on the memory expands the sample diversity, and also balances the number to a small extent. However, the randomness of DA still limits its effectiveness on OCI. For the purpose of adjusting the decision boundary more directly, we induce DA to generate samples near the decision boundaries by mixing samples from old and new classes. In order to push the decision boundary farther away from old classes , we adaptively adjust the mixing ratio on labels by AdpMix:
\begin{equation}
    \begin{aligned}
        & x^a=\mu _x x_i+\left( 1-\mu _x\right) x_j \\
        & y^e=\mu _y y_i+\left( 1-\mu _y\right) y_j
    \end{aligned}
    \label{eq: adpmix}
\end{equation}
where $x_i$ and $x_j$ are from old classes (memory data) and new classes (current task data). Equation~\ref{eq: adpmix} is similar to Equation~\ref{eq: enmix}. The difference are two-fold: 1) EnMix mixes the augmented samples from memory and AdpMix mixes the original data from memory and imminent task dataset. 2) EnMix use the same ratio $\mu$ for raw data and label while the ratio $\mu _y$ for label is different from the ratio $\mu _x$ for data in AdpMix. The main idea is that when the decision boundary is too closer to old classes, the mixing ratio $\mu _y$ should be also more biased towards old classes, i.e., giving $\mu _y$ a larger value relative to $\mu _x$.

The main reason for class imbalance is from the quantity problem, but in OCI, we cannot know the number of new task samples in advance, making it inappropriate to derive the value of $\mu$ based on the quantity gap. In addition, CL retains the knowledge stored in memory by learning from previous tasks. Therefore, it is not reliable to evaluate the imbalance only by the sample size. A recent work~\cite{ahn2020simple} revealed that the class imbalance results in a biased weights $\mathbf{w}$ of linear classifier. Therefore, we use the weights $\mathbf{w}$ to design the adjustment scheme of $\mu _y$:
\begin{equation}
\begin{split}
\mu _y=\left\{ \begin{matrix}
	\min \left( \mu _x+\delta \frac{\left\| \mathbf{w}_{new} \right\|}{\left\| \mathbf{w}_{old} \right\|},1 \right) ,&		\frac{\left\| \mathbf{w}_{new} \right\|}{\left\| \mathbf{w}_{old} \right\|}>\kappa ,\mu _x>\tau \,\,\\
	\mu _x,&		others\\
\end{matrix} \right. \,\,
\end{split}
\label{eq: mu_y adjust}
\end{equation}
where $old$ or $new$ denotes an index for old or new class respectively. $\delta$, $\kappa$ and $\tau$ are the hyper-parameters. $\kappa$ is used to confirm the occurrence of class imbalance. $\delta $ and $\tau$ control the degree of the adaptive adjustment. When the decision boundary deviation occurs, we push it away from old classes. The magnitude of the adjustment is calculated according to the relative L-2 norm value of weights.

We rewrite the output probability of class $c$ to show how AdpMix can adjust the updating of weights $\mathbf{w}$:
\begin{equation}
    p_{i,c}=\frac{\exp \left( \mathbf{w}_{c}^{T}\mathbf{h}_i \right)}{\sum\nolimits_{j=1}^C{\exp \left( \mathbf{w}_{c}^{T}\mathbf{h}_j \right)}}
    \label{eq: softmax prob}
\end{equation}

For simplicity, we assume that the buffer is not updated during the training of current task $t$, that is, the previous tasks data and current task data are clearly divided into $D^\mathcal{M}$ and $D_t$. The CE loss is calculated as:
\begin{equation}
    L=-\sum_{x_i\in D_t\cup D^{\mathcal{M}}}{\sum_{c=1}^C{\mathcal{I} \left\{ y_i=c \right\} \log p_{i,c}}}
    \label{eq: CE loss}
\end{equation}
where $\mathcal{I} \left\{ \cdot \right\}$ is the indication function. According to gradient calculation of CE loss, we can obtain the update formula of $\mathbf{w}_c$ as:
\begin{equation}
    \begin{split}
        \mathbf{w}_c=\mathbf{w}_c+\underset{current\,\,task\,\,data}{\underbrace{\eta \sum_{x_i\in D_t,y_i=c}{\left( 1-p_{i,c} \right) \mathbf{h}_i}-\eta \sum_{x_i\in D_t,y_i\ne c}{p_{i,c}\mathbf{h}_i}}}
\\
+\underset{memory\,\,data}{\underbrace{\eta \sum_{x_i\in D^{\mathcal{M}},y_i=c}{\left( 1-p_{i,c} \right) \mathbf{h}_i}-\eta \sum_{x_i\in D^{\mathcal{M}},y_i\ne c}{p_{i,c}\mathbf{h}_i}}}
    \end{split}
    \label{eq: CE gradient}
\end{equation}

\begin{table*}
    \renewcommand\arraystretch{1.1}
    \begin{center}
    \setlength{\tabcolsep}{1.45mm}{
            \begin{tabular}{c|c c c|c c c|c c c}
        \hline
        \hline
        Method & \multicolumn{3}{c|}{CIFAR-100} & \multicolumn{3}{c|}{Mini-ImageNet} & \multicolumn{3}{c}{Tiny-ImageNet} \\
        \hline
        Finetune & \multicolumn{3}{c|}{3.4 ± 0.4} & \multicolumn{3}{c|}{3.2 ± 0.2} & \multicolumn{3}{c}{2.4 ± 0.5} \\
        iid offline & \multicolumn{3}{c|}{48.8 ± 2.1} & \multicolumn{3}{c|}{52.1 ± 1.0} & \multicolumn{3}{c}{39.4 ± 2.2} \\
        \hline
        Memory Size & M=1K & M=2K & M=5K & M=1K & M=2K & M=5K & M=1K & M=2K & M=5K \\
        \hline
        A-GEM & 3.2 ± 0.3 & 3.4 ± 0.2 & 3.2 ± 0.3 & 2.9 ± 0.3 & 3.2 ± 0.3 & 3.7 ± 0.3 & 2.6 ± 0.2 & 2.5 ± 0.2 & 2.6 ± 0.3 \\
        SCR & 14.3 ± 0.8 & 16.6 ± 0.5 & 17.5 ± 0.4 & 13.1 ± 0.4 & 14.8 ± 0.6 & 15.9 ± 0.6 & 6.2 ± 0.7 & 7.8 ± 0.4 & 9.4 ± 0.6 \\
        DVC & 16.3 ± 0.7 & 17.8 ± 0.9 & 20.2 ± 1.5 & \textbf{14.0 ± 0.8} & 16.4 ± 2.0 & 16.8 ± 1.1 & 6.8 ± 1.2 & 8.4 ± 1.4 & 12.0 ± 1.2 \\
        \hline
        ER & 8.4 ± 0.5 & 12.7 ± 0.8 & 18.0 ± 1.1 & 7.8 ± 0.7 & 11.9 ± 1.2 & 15.0 ± 2.6 & 4.6 ± 0.4 & 6.3 ± 0.6 & 10.4 ± 0.8 \\
        ER+DualMix & \textbf{17.5 ± 1.0} & \textbf{20.3 ± 0.9} & \textbf{22.5 ± 1.2} & 13.2 ± 0.8 & \textbf{16.7 ± 0.8} & \textbf{17.9 ± 1.1} & \textbf{6.9 ± 0.9} & 9.4 ± 1.4 & 13.6 ± 1.1 \\
        \hline
        GSS & 7.9 ± 0.6 & 11.1 ± 0.9 & 15.1 ± 0.9 & 7.7 ± 0.9 & 11.5 ± 0.9 & 12.8 ± 1.0 & 4.1 ± 0.4 & 5.7 ± 0.7 & 9.7 ± 1.0 \\
        GSS+DualMix & 9.4 ± 0.8 & 13.4 ± 1.1 & 15.7 ± 1.5 & 9.5 ± 1.2 & 12.3 ± 1.5 & 13.7 ± 0.9 & 5.5 ± 0.8 & 6.4 ± 1.0 & 10.9 ± 0.9 \\
        \hline
        MIR & 8.7 ± 0.4 & 13.2 ± 0.9 & 18.6 ± 1.2 & 8.0 ± 0.6 & 12.8 ± 1.1 & 16.5 ± 2.2 & 4.2 ± 0.5 & 6.0 ± 0.5 & 11.3 ± 1.0 \\
        MIR+DualMix & 17.8 ± 0.8 & 17.2 ± 2.3 & 19.2 ± 2.8 & 11.1 ± 1.3 & 14.8 ± 1.8 & 17.1 ± 2.7 & 7.1 ± 1.2 & \textbf{9.9 ± 0.8} & \textbf{13.9 ± 0.4} \\
        \hline
        ASER & 13.1 ± 0.7 & 16.4 ± 0.6 & 21.0± 0.8 & 8.9 ± 1.3 & 11.7 ± 0.9 & 16.3 ± 2.1 & 7.1 ± 0.3 & 8.8 ± 0.9 & 11.6 ± 0.7 \\
        ASER+DualMix & 13.8 ± 0.8 & 17.9 ± 2.4 & 21.7± 1.5 & 9.6 ± 0.9 & 13.1 ± 1.8 & 16.9 ± 1.9 & 6.3 ± 0.5 & 9.5 ± 0.6 & 12.2 ± 1.0 \\
        \hline
        \hline
        \end{tabular}
    }
    \end{center}
    \caption{Average Accuracy (end of training, higher is better) on three benchmarks with 1K, 2K and 5K memory. DualMix boosts the performance of ER-based baselines, making them even better than pervious SOTA (DVC) in most cases.}
    \label{tab: Accuracy comparison with baseline}
\end{table*}

Because the activation function for the feature extractor \cite{he2016deep} is usually ReLU, $\mathbf{h}_i$ is always positive. Due to the number of samples in $D_t$ is usually far more than that of each class samples in $D^\mathcal{M}$, the gain or reduction for $\mathbf{w}_c$ is mainly affected by the current task $D_t$. We can see that the weights of class $c$ from old tasks are extremely weakened as no sample with class $c$ in $D_t$ (the first term in \textit{current task data} equals to 0), while the weights of class from current task are going to be effectively increased. When we apply AdpMix as Equation~\ref{eq: adpmix}, Equation~\ref{eq: CE gradient} will be changed to (the \textit{memory data} term is omitted because of lesser influence):
\begin{equation}
    \begin{split}
        \mathbf{w}_c=\mathbf{w}_c+\eta \sum_{x_i\in D_t,y_i=c}{\left( \mu _y -p_{i,c} \right) \mathbf{h}_i}-\eta \sum_{x_i\in D_t,y_i\ne c}{p_{i,c}\mathbf{h}_i}
    \end{split}
    \label{eq: mix gradient}
\end{equation}
When $c$ belongs to the current task, the ratio $\mu _y$ reduces the excessive growth of the corresponding weights. When $c$ is a old class, the mixture operation makes the gain not equal to 0. And we further balance the weights by adaptively adjusting $\mu _y$. This also shows that it is reasonable to adjust the mixing ratio according to the weights of the classifier.



\section{Evaluation}
\subsection{Datasets}
\noindent\textbf{Split CIFAR-100} is constructed by splitting the CIFAR-100 dataset \cite{krizhevsky2009learning} into 20 tasks with no overlapping classes between each task. Every task is randomly assigned 5 classes data. The image size is $3\times 32\times 32$. There are a total 3,000 images in each class, which are divided into 2,500 training samples and 500 test samples.

\noindent\textbf{Split Mini-ImageNet} splits the Mini-ImageNet dataset \cite{vinyals2016matching}, containing 100 classes, into 20 disjoint tasks as in~\cite{mai2022online}. Each task includes 5 random classes, and every class consists of 500 $3\times 84\times 84$ images for training and 100 images for testing.

\noindent\textbf{Split Tiny-ImageNet} is used to verify algorithms effectiveness in more complex scenarios. We split it Tiny-ImageNet \cite{le2015tiny} into 20 disjoint tasks and each task contains 10 classes. Each class contains 500 $3\times 64\times 64$ images for training and 100 images for testing.


\subsection{Baselines and metrics}
The baselines we compare are the methods mentioned in Section \ref{sec: related}: A-GEM, and ER, GSS, MIR, ASER, focusing on sample selection, and SCR and DVC, two algorithms with standard DA.
We apply our DualMix on the four ER-based methods and compare them with A-GEM, SCR and DVC. Finetune and iif offline are used as the lower and upper bounds as the same setting in \cite{mai2022online}.

We use two standard metrics in the continual learning to measure performance: Average Accuracy and Average Forgetting. Let $a_{i,j}$ be the accuracy of the model on testing set of task $j$ after trained from task 1 to task $i$. $f_{i,j}$ represents how much the model has forgot about task $j$ after being
trained on task $i$. The two metrics are defined as:
\begin{equation}
    \mathrm{Average}\, \, \mathrm{Accuracy}\left( A_i \right) =\frac{1}{i}\sum_{j=1}^i{a_{i,j}}
    \label{eq: average accuracy}
\end{equation}
\begin{equation}
    \begin{split}
        \mathrm{Average}\, \, \mathrm{Forgetting}\left( F_i \right) =\frac{1}{i-1}\sum_{j=1}^{i-1}{f_{i,j}}
\\
\mathrm{where}\, \, f_{k,j}=\underset{l\in \left\{ 1,\cdots ,k-1 \right\}}{\max}\left( a_{l,j} \right) -a_{k,j},\forall j<k
    \end{split}
    \label{eq: average forgetting}
\end{equation}

\begin{table*}
    \renewcommand\arraystretch{1.1}
    \begin{center}
    \setlength{\tabcolsep}{1.45mm}{
        \begin{tabular}{c|c c c|c c c|c c c}
        \hline
        \hline
        Dataset & \multicolumn{3}{c|}{CIFAR-100} & \multicolumn{3}{c|}{Mini-ImageNet} & \multicolumn{3}{c}{Tiny-ImageNet} \\
        \hline
        Memory Size & M=1K & M=2K & M=5K & M=1K & M=2K & M=5K & M=1K & M=2K & M=5K \\
        \hline
        A-GEM & 57.4 ± 0.8 & 58.0 ± 1.1 & 57.3 ± 1.2 & 52.3 ± 1.2 & 52.1 ± 1.3 & 52.5 ± 1.5 & 44.4 ± 1.0 & 43.8 ± 1.1 & 44.5 ± 0.6 \\
        ER & 52.1 ± 1.5 & 47.9 ± 1.1 & 44.4 ± 1.1 & 46.8 ± 1.8 & 43.6 ± 1.8 & 42.1 ± 1.6 & 45.6 ± 1.5 & 43.0 ± 1.4 & 39.1 ± 1.6 \\
        GSS & 51.1 ± 1.1 & 47.5 ± 1.1 & 44.5 ± 1.8 & 47.4 ± 1.8 & 41.6 ± 1.4 & 39.4 ± 1.3 & 45.1 ± 1.0 & 42.9 ± 1.2 & 41.1 ± 1.4 \\
        MIR & 50.2 ± 1.0 & 42.9 ± 1.7 & 40.6 ± 1.5 & 44.9 ± 1.1 & 39.8 ± 1.6 & 37.4 ± 2.1 & 44.9 ± 2.0 & 40.9 ± 1.8 & 36.7 ± 1.9 \\
        ASER & 52.1 ± 1.0 & 46.6 ± 0.8 & 37.9 ± 1.4 & 48.1 ± 1.4 & 44.8 ± 1.3 & 38.7 ± 2.2 & 46.4 ± 0.8 & 43.1 ± 0.6 & 39.7 ± 0.4 \\
        DVC & 40.2 ± 1.0 & 38.8 ± 1.1 & 37.2 ± 1.4 & 40.0 ± 1.0 & 37.0 ± 1.7 & 36.6 ± 2.4 & 40.5 ± 1.6 & 36.7 ± 1.7 & 33.2 ± 3.1 \\
        \hline
        ER+DualMix & \textbf{36.7 ± 1.4} & \textbf{31.8 ± 1.7} & \textbf{32.8 ± 1.1} & \textbf{35.7 ± 1.1} & \textbf{31.2 ± 1.7} & \textbf{30.4 ± 2.1} & \textbf{29.5 ± 2.1} & \textbf{22.7 ± 2.0} & \textbf{26.5 ± 1.9} \\
        \hline
        \hline
        \end{tabular}
        }
    \end{center}
    \caption{Average Forgetting (end of training, lower is better) on three benchmarks with 1K, 2K and 5K memory. DualMix is only applied on ER, which is enough to demonstrate our good performance. }
    \label{tab: Forgetting comparison with baseline}
\end{table*}

\subsection{Implementation details}
We use a reduced ResNet-18 for all datasets as in \cite{mai2022online, gu2022not}. A single-head is used for all classes. The normal DA used here is a combination of four augmentation methods as used in \cite{gu2022not, mai2021supervised}: random crop, horizontal flip, color jittering and grayscale. We use Stochastic Gradient Descent (SGD) to optimize the model and set the learning rate to 0.1, batch size to 10. $\alpha$ is set to 0.2. $\kappa$, $\tau$ and $\delta$ are set to 2.0, 0.5 and 0.05 respectively. EnMix and AdpMix are both combined with standard DA. All the experimental results we present are averages of 10 runs, performed on one NVIDIA GeForce RTX 3090 GPU.

\subsection{Comparative performance evaluation}
\noindent\textbf{Accuracy performance.} The accuracy results are illustrated in Table \ref{tab: Accuracy comparison with baseline}. We apply our DualMix strategy on four ER-based methods, ER, GSS, MIR and ASER, which optimize the sample selection for memory in OCI. According to the results, DualMix can improve the ER-based methods significantly in the three datasets, showing that our mixing method is effective on a variety of sample distributions. As we can see, our method performs better in relatively small memory (double the accuracy on CIFAR-100 with ER and MIR), due to the severe lack of data diversity in such setting. In Mini-ImageNet, our method can also achieve 69.2\% and 40.3\% performance gain with 1K and 2K memory respectively. Interestingly, compared with ER, the gains on other ER-based methods are relatively small. Intuitively, the combination of better methods should lead to better performance. However, our method increases the performance on ASER and GSS with a limited degree. On the one hand, this may be due to the inherently advanced performance of these method, making it difficult to keep improving. On the other hand, sample selection strategy produces a distribution that is quite different from the original data, resulting in biased augmentation. In contrast, the samples obtained by ER through reservoir sampling \cite{vitter1985random} are more uniform. Combined with DA, there is a greater potential to obtain more diverse and adequate samples. In addition to effectively improving the performance of some existing methods, our strategy also demonstrates superiority over state-of-the-art methods, SCR and DVC. For a fair comparison, the memory batch size of SCR is the same with our DualMix and DVC. It can be seen that our method performs remarkably better than DVC in most cases, despite the fact that DVC and SCR still require new loss functions to enhance feature exploration about old and current tasks.
These results suggest that DA plays an important role in OCI, and its potential has not been fully tapped in previous approaches.

\noindent\textbf{Forgetting rate.} Table \ref{tab: Forgetting comparison with baseline} shows the Average Forgetting by the end of training. We apply DualMix on ER to compare with other baselines in CIFAR-100, Mini-ImageNet and Tiny-ImageNet.
We don't illustrate the Average Forgetting results of SCR because it performs poorly on tasks due to limited memory batch size.
Our method can achieve the lowest forgetting on the three benchmark datasets with different memory sizes. In CIFAR-100, our method can achieve 8.7\% $\sim $ 18.0\% reduction on Average Forgetting compared with the strongest baseline DVC. This number in Tiny-ImageNet can even reach 38.1\% with 2K memory, opening up a huge gap with other methods. 

\begin{figure}
    \centering
    \includegraphics[width=0.75\linewidth]{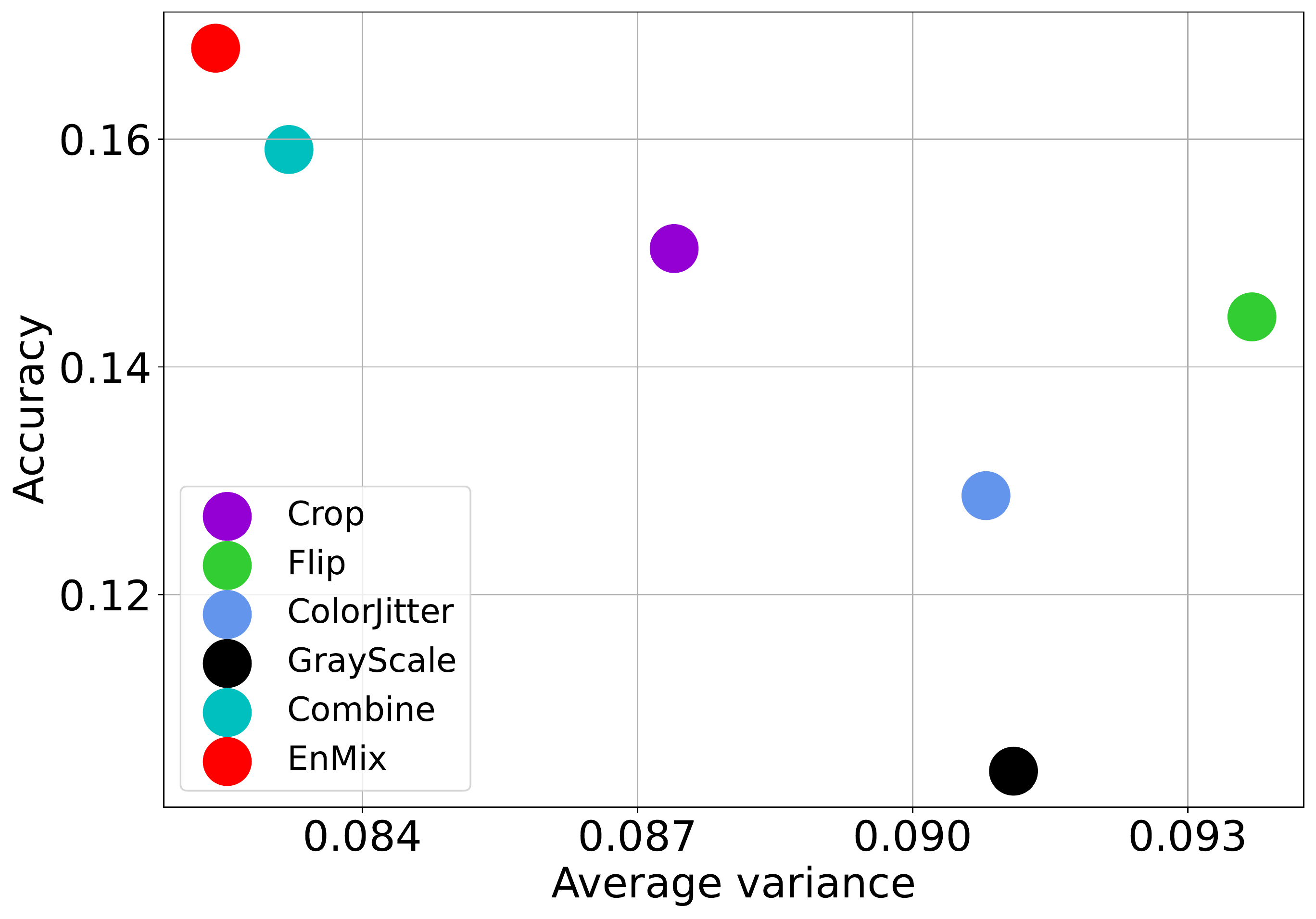}
    \caption{The mean variance calculated by Equation \ref{eq: corr calculation} on CIFAR-100 with different DA methods. ``Combine" means the combination of ``Crop", ``Flip", ``ColorJitter" and ``GrayScale" and EnMix is performed based on ``Combine" to get stronger DA.}
    \label{fig: average variance}
\end{figure}

\subsection{Ablation study}
\label{sec: ablation}
\noindent\textbf{Effectiveness of each component.} We investigate the effectiveness of each component of our method. As shown in Table \ref{tab: Ablation}, a strong DA can already make improvement compared with the baseline. Applied the stronger EnMix, further great improvement occurs on Average Accuracy and Average Forgetting in all cases. In addition, AdpMix also improves the performance of the model in various experiments, which is even more remarkable than EnMix. After combining all the components, our method achieves the best performance especially when the memory is small. This is an intuitive phenomenon, which highlights the importance of data augmentation to enrich diversity and alleviate the class imbalance. The results indicate that each of our components is essential.

\begin{table*}
    \renewcommand\arraystretch{1.1}
    \begin{center}
    \setlength{\tabcolsep}{6mm}{
        \begin{tabular}{c|c|c|c}
        \hline
        \hline
        Method & M=1k (AA$\uparrow$/AF$\downarrow$) & M=2k (AA$\uparrow$/AF$\downarrow$) & M=5k (AA$\uparrow$/AF$\downarrow$) \\
        \hline
        ER & 8.4 ± 0.5 / 52.1 ± 1.5 & 12.7 ± 0.8 / 47.9 ± 1.1 & 18.1 ± 1.1 / 44.4 ± 1.1  \\
        ER+DA & 14.6 ± 1.6 / 48.0 ± 2.4 & 15.5 ± 2.0 / 48.5 ± 2.6 & 18.5 ± 1.4 / 43.2 ± 1.9  \\
        ER+EnMix & 15.7 ± 1.1 / 42.7 ± 1.6 & 17.2 ± 4.0 / 43.4 ± 3.9 & 18.7 ± 1.4 / 43.8 ± 2.3  \\
        ER+AdpMix & 16.6 ± 1.2 / 38.7 ± 1.8 & 18.9 ± 1.8 / 33.5 ± 3.8 & 21.3 ± 2.2 / 33.3 ± 2.4  \\
        \hline
        ER+DualMix & \textbf{17.5 ± 1.0 / 36.7 ± 1.4} & \textbf{20.3 ± 0.9 / 31.8 ± 1.7} & \textbf{22.5 ± 1.2 / 32.8 ± 1.1}  \\
        \hline
        \hline
        \end{tabular}
        }
    \end{center}
    \caption{Comparison with different augmentation strategies on CIFAR-100 with 1K, 2K and 5K memory. AA and AF denote Average Accuracy and Average Forgetting respectively. Every component is essential in our method and their combination achieves the best performance.}
    \label{tab: Ablation}
\end{table*}

\noindent\textbf{Empirical results about EnMix and correlation.} To verify our proposition 1, we use the mean variance of the output probability based on old models, as described in Equation \ref{eq: corr calculation}, to demonstrate the effect of EnMix. 
In Figure~\ref{fig: average variance}, 
we plot the scatters of $\bar{m}$ and the final Average Accuracy on CIFAR-100 with various DA strategies. 
We can see that there exists an obviously negative correlation between covariance and accuracy.
``Flip", ``Colorjitter" and ``GrayScale" are with relatively weak strength, so the augmentation samples have higher correlation with original samples, which results in weaker improvement. The strength of ``Crop" and their combination is stronger and our EnMix further promotes them, achieving better performance. 

\begin{figure}
    \centering
    \includegraphics[width=0.77\linewidth]{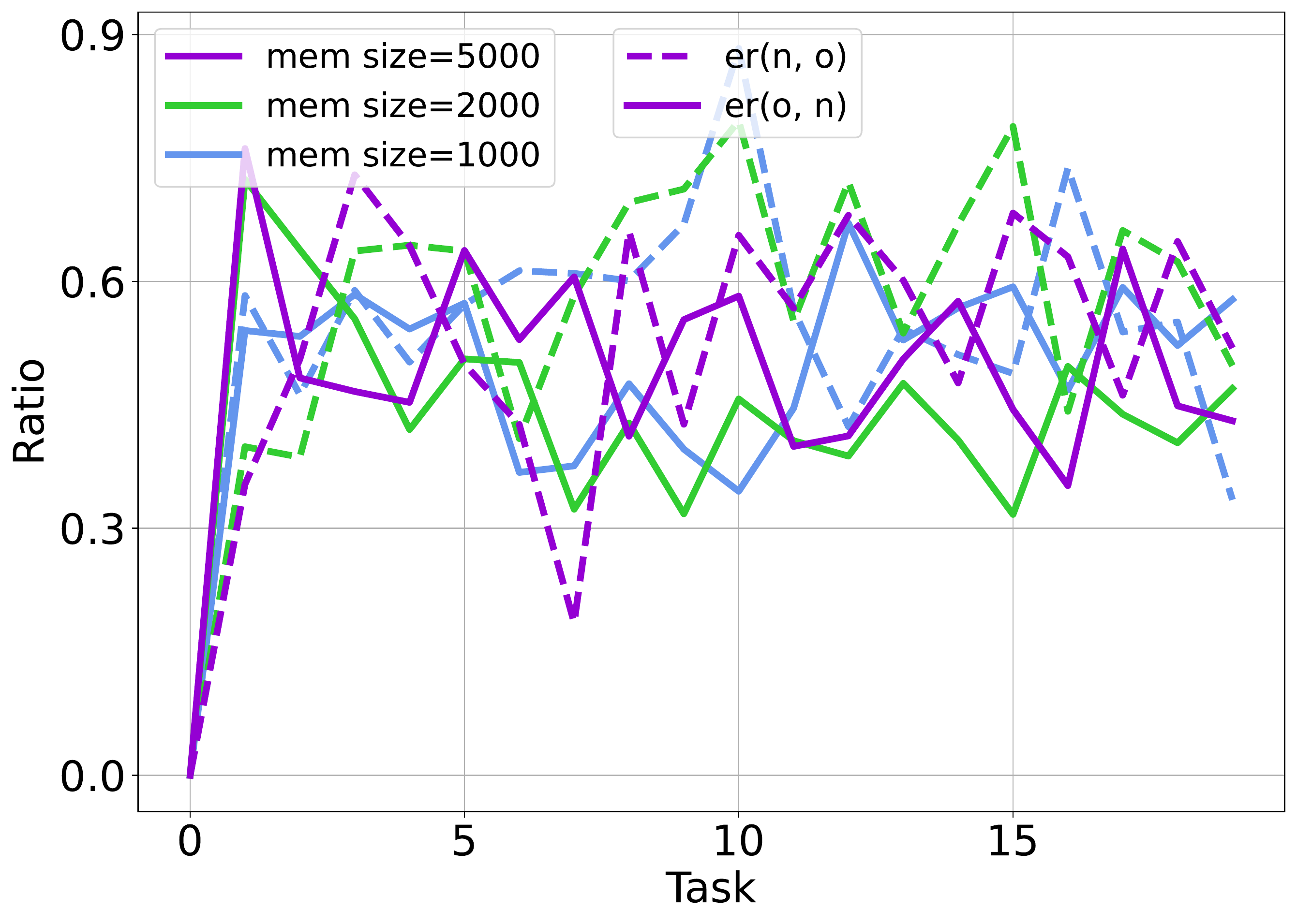}
    \caption{The misclassification ratios during the training process on CIFAR-100 by ER+AdpMix with different memory sizes. The biased decision boundary is greatly alleviated.}
    \label{fig: AdpMix-ratio}
\end{figure}

\noindent\textbf{Correct decision boundary with AdpMix.} As we discussed in \ref{sec: insight}, there exists the serious biased decision boundary between old and new classes. Therefore, we propose the AdpMix to push the decision boundary away from samples of old classes. As shown in Figure \ref{fig: AdpMix-ratio}, we plot the error ratio of $er\left( n,o \right)$ and $er\left( o,n \right)$ when we apply AdpMix on ER. Compared with the results in Figure \ref{fig: ER-ratio}, $er\left( o,n \right)$ decreases notably and the gap between $er\left( o,n \right)$ and $er\left( n,o \right)$ is almost wiped out. This shows that our method does correct the previous decision boundary.

\begin{figure}
    \centering
    \includegraphics[width=0.82\linewidth]{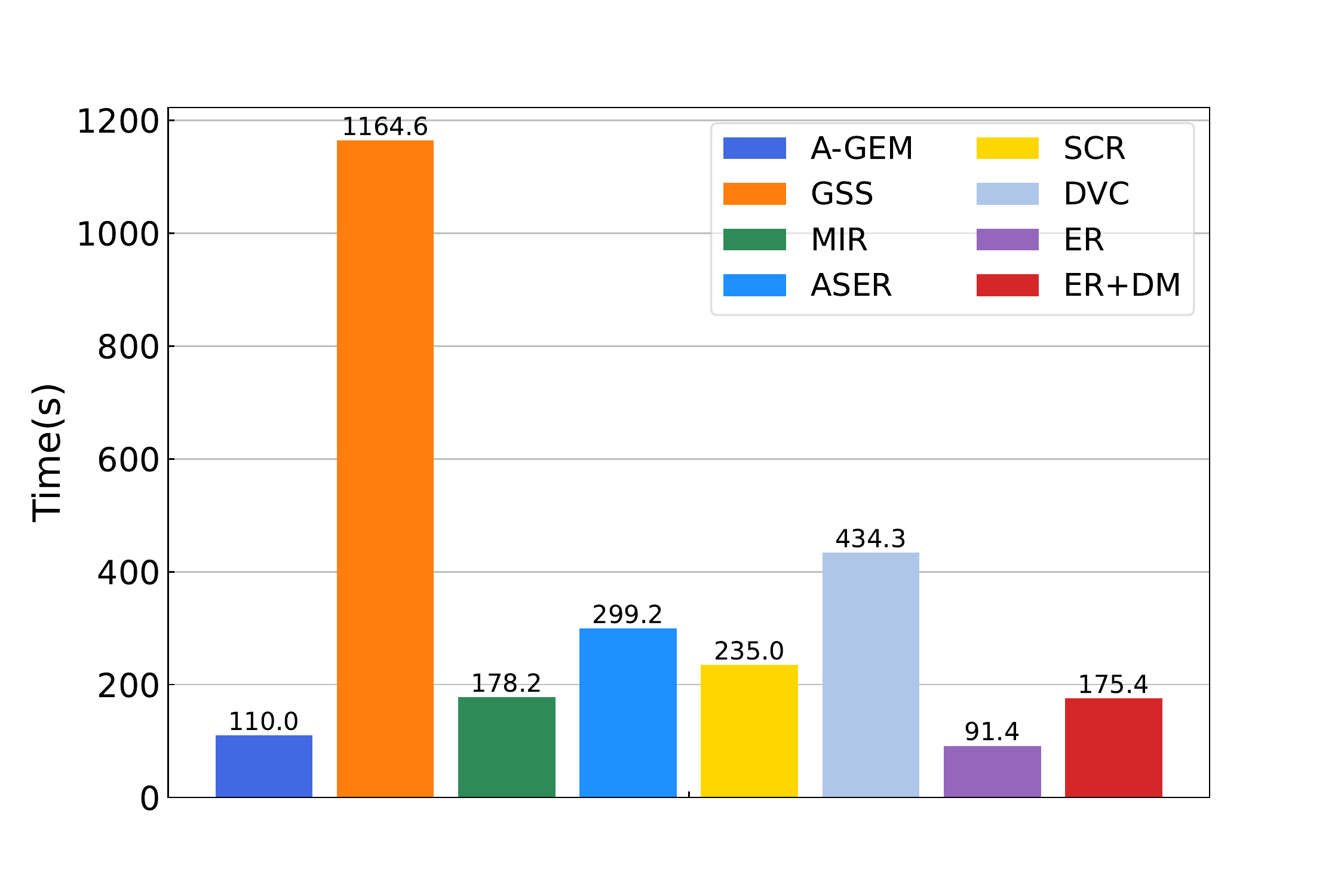}
    \caption{Running time comparison with a 2K memory buffer and trained on CIFAR-100. }
    \label{fig: time comparison}
\end{figure}

\noindent\textbf{Running time comparison.} Our method only perform mixing augmentation to enrich samples, which adds a certain amount of training time. However, the increased time cost of our method is pretty small than that of some other sample selection or contrastive-relevant methods. We can see from Figure \ref{fig: time comparison} that ER and A-GEM, which use the reservoir method to update and retrieve memory, have the shortest training time. Various sample selection strategies are applied in other approaches, significantly increasing running time especially SCR, DVC and GSS. Our method only augments memory data, and the increased running time is less and within a controllable range. 
\section{Discussion}
Continual learning can falls into the poor performance because of catastrophic forgetting. We revisit the data augmentation strategy in online class-incremental learning, a challenging and more realistic setting. We prove that aggressive augmentation which generates samples with low correlation with original samples are beneficial to forgetting avoidance. However, label consistency occurs when the standard augmentation is too strong. We propose the Enhanced Mixup (EnMix), which mixes the augmented samples and their labels based on standard DA, resulting in increased sample diversity and consistency with labels. Further, to address the class imbalance in replay-based OCI, we introduce the Adaptive Mixup (AdpMix) to mix samples from old and new classes, which can recalibrate the biased decision boundary. Our method can be directly combined with existing methods and boost them effectively. The practice in this paper shows that the CF in CL can be greatly alleviated merely through appropriate DA strategy.

{\small
\bibliographystyle{ieee_fullname}
\bibliography{egbib}
}

\end{document}